\documentclass{article}
\usepackage{ijcai16}

\usepackage{times}

\usepackage{graphicx}
\usepackage{multirow}
\usepackage{booktabs}

\title{Helping News Editors Write Better Headlines: \\ \large A Recommender to Improve the Keyword Contents \& Shareability of News Headlines}

\author{Terrence Szymanski, Claudia Orellana-Rodriguez, Mark T. Keane \\ 
Insight Centre for Data Analytics \&\\
School of Computer Science \\
University College Dublin\\
\{terrence.szymanski,claudia.orellana,mark.keane\}@insight-centre.org}

\begin{document}

\maketitle

\begin{abstract}
We present a software tool that employs state-of-the-art natural language processing (NLP) and machine learning techniques to help newspaper editors compose effective headlines for online publication. The system identifies the most salient keywords in a news article and ranks them based on both their overall popularity and their direct relevance to the article. The system also uses a supervised regression model to identify headlines that are likely to be widely shared on social media. The user interface is designed to simplify and speed the editor's decision process on the composition of the headline. As such, the tool provides an efficient way to combine the benefits of automated predictors of engagement and search-engine optimization (SEO) with human judgments of overall headline quality.
\end{abstract}


\section{Introduction}

The headline is an extremely important component of every news article that performs multiple functions: summarizing the story, attracting attention, and signaling the voice and style of the newspaper \cite{Conboy:2007}. In the online realm, headlines are expected to meet several new functions; for instance, to convey the article's contents in different online contexts or to optimize the article for search engine queries (i.e, SEO).  Indeed, arguably, the headline is now more important than ever, as it  becomes the only visible part of the article in microblog posts, social media feeds and listings on news-aggregation sites. These multiple requirements on the news headline have complicated the composition task facing news editors, as they attempt to ensure that each headline is crafted as perfectly as possible.

Prior NLP work in the area of news headlines has mostly focused on the task of automatic headline generation, cast as ``very short summary generation'' in the DUC tasks of the early 2000s; tasks that produced much of the research on the topic. The best-performing system in the 2004 DUC task worked by parsing the first sentence of the article and pruning it to the desired length \cite{Zajic:2004}, an approach that works by leveraging human intelligence: journalists generally compose news articles in the ``inverted pyramid'' style, which places the most important information  in the lead paragraph \cite{Conboy:2007}. Other headline generation systems generally work by first using some metric to identify terms within the document that are likely to appear in the headline, and then constructing a headline containing these terms \cite{Nenkova:2011}.

This latter approach has much in common with the task of keyword selection for SEO, which first caught the attention of major newspapers at least ten years ago \cite{Lohr:2006}, and continues to be a much-discussed issue today \cite{Sullivan:2015}. While even long-established, traditional news publications have begun to move away from classical forms of headlines towards more direct, keyword-laden headlines, many copy editors would still prefer to write clever, witty headlines \cite{Wheeler:2011}, and readers of the news seem to value creativity in headlines over clarity or informativeness \cite{Ifantidou:2009}. Therefore, one of the key considerations in the design of our system was to balance the mechanical act of filling a headline with informative, relevant keywords, against the creative act of writing headlines that appeal to human interests and emotions.

We expect that the most interesting and emotional stories are likely to be more popular with readers than the ``average'' story. Analysis of reader behavior has shown that there is no correlation between how much an article is shared on social media and how much of the article is read by an average user \cite{Haile:2014}; a fact that could be taken as evidence supporting the widely-held view that people share articles online that they have not fully read themselves \cite{Manjoo:2013}. In this case, the headline---which people presumably read even if they don't read the full text---may be an important factor in determining the ``shareability'' of a news article; an idea that is another key motivation behind the design of our system.

The tool presented here is designed to facilitate the decision-making process facing a news editor in composing a headline. The software employs state-of-the-art NLP and machine learning techniques to make its recommendations, but it is not designed to automatically generate headlines or to make decisions about a headline's goodness on its own. 

In the sections below, we present the design and behavior of the tool before discussing the internal workings of the system. We conclude with an assessment of the current state of the project, including some preliminary evaluation results and a discussion of areas for improvement. 


\section{Design and Behavior}
\label{sec:Design}

\begin{figure}[t]
  \centering
  \includegraphics[width=3.4in]{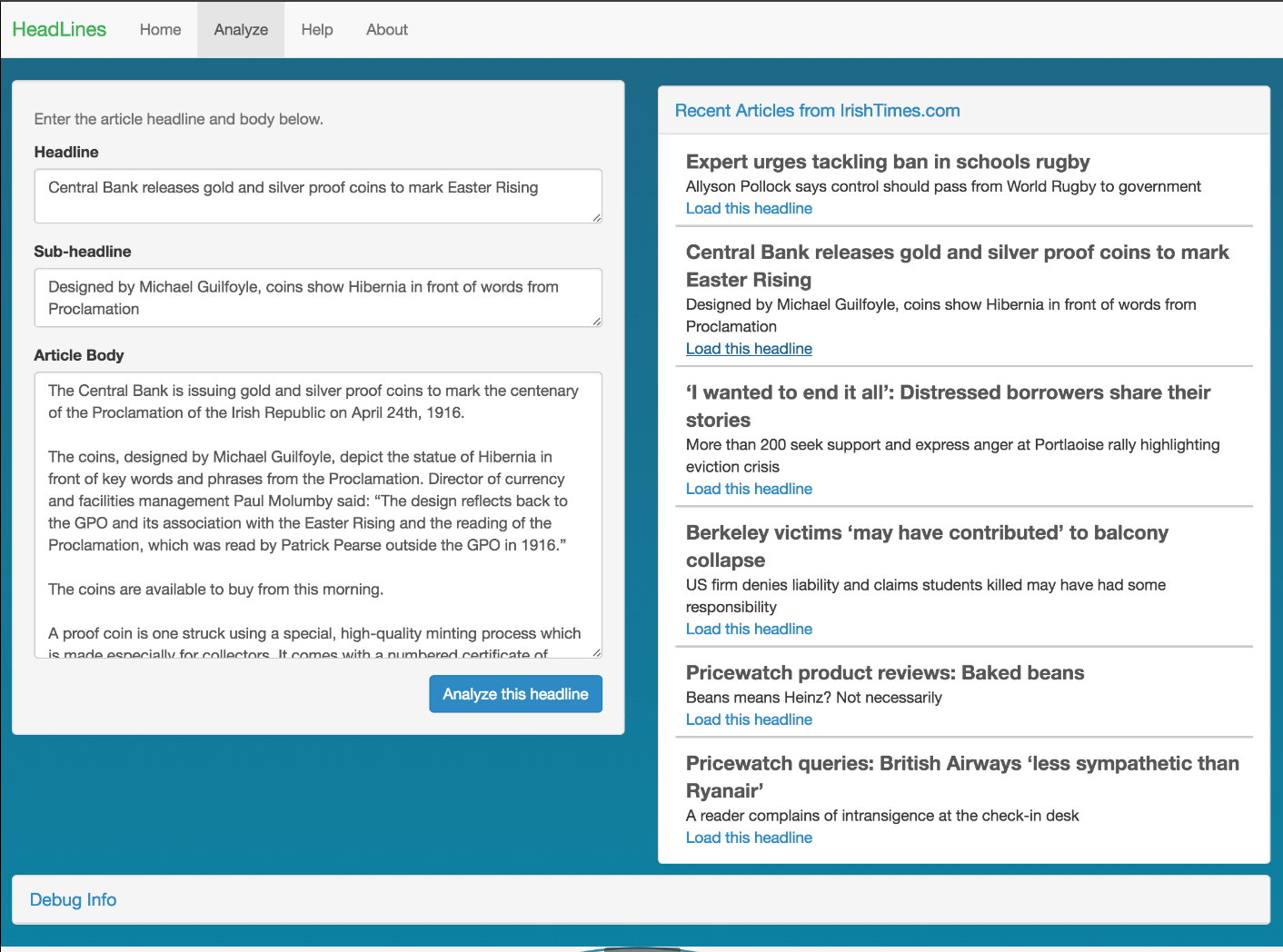}
  \caption{Screenshot of the tool in input mode, with input text areas on the left and a live feed of articles on the right.}
  \label{fig:screenshot1}
\end{figure}


From a user-interface perspective, the software has two modes of operation: input mode and analysis mode. The input mode (illustrated in Figure~\ref{fig:screenshot1}) facilitates the entry of a news article and its corresponding headline and sub-headline, which may either be entered manually or selected from a feed of recent articles. In practice, this feed would be integrated into the newspaper's workflow so that an editor could review all new articles with the software prior to publication.

\begin{figure*}[t]
  \centering
  \includegraphics[width=7in]{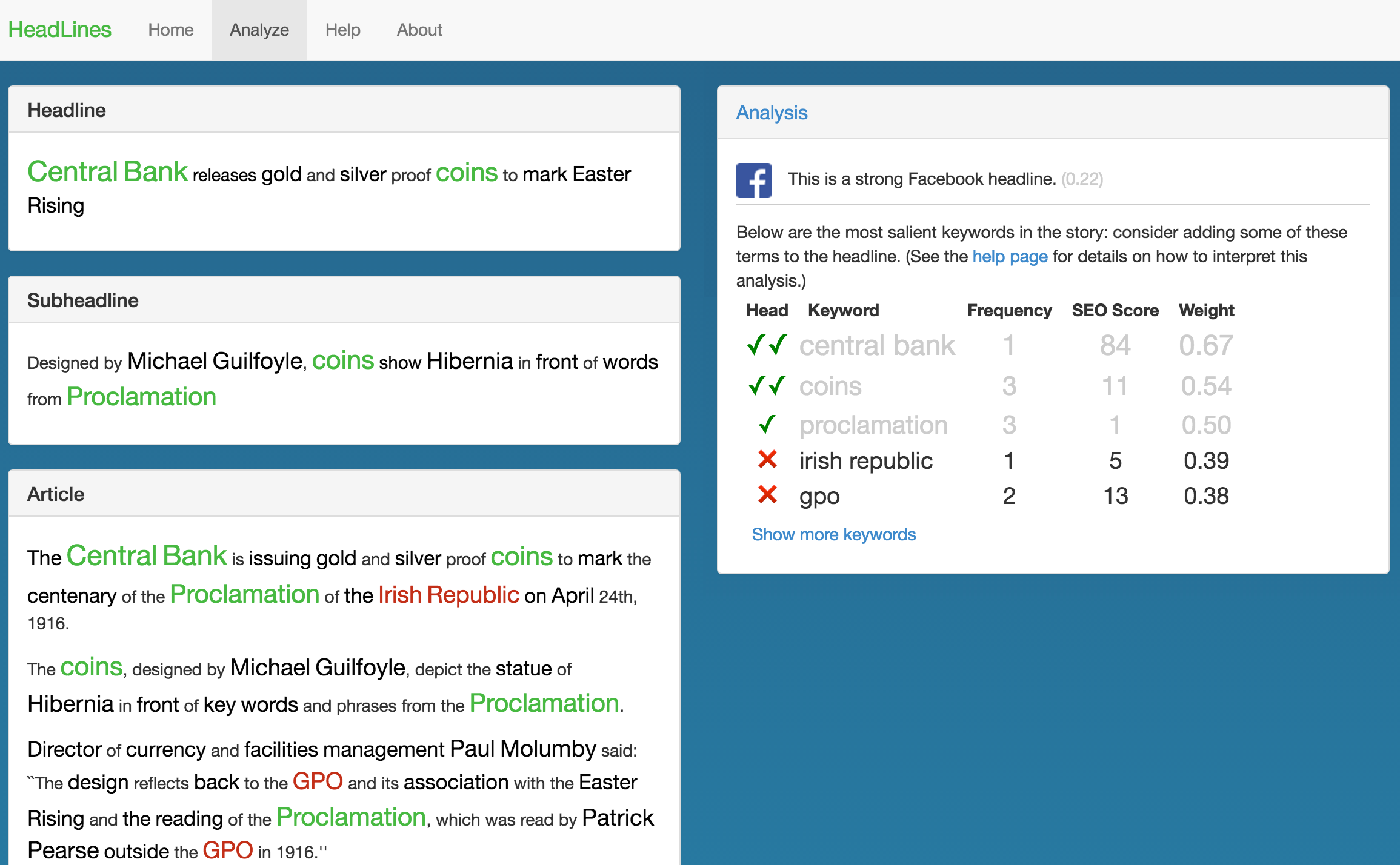}
  \caption{Screenshot of the tool in analysis mode. In this example, three of the top five keywords (highlighted in green) are already in the headline;  the remaining two (in red) are recommendations that the user may consider adding to the headline.
}
  \label{fig:screenshot3}
\end{figure*}

After the editor-user selects an article, the system switches to the analysis mode, showing the results of the automated analysis (illustrated in Figure~\ref{fig:screenshot3}). This mode is designed to allow the user to quickly assess the strengths and weaknesses of the headline and decide whether any changes should be made to improve it. The five most highly-ranked keywords from the article are listed on the right side of the screen sorted by {\em weight}, a metric combining the keyword's {\em frequency} in the article and its {\em SEO score}, which respectively capture the keyword's {local} relevance to the article itself as well as its {global} prominence among news stories in general. (See section~\ref{sec:Keyword} below for details on these measures.)

The keywords are color-coded to distinguish keywords which already appear in the headline (green) from those which do not appear in the headline (red), and size-coded according to their weight. Thus, any large, red keywords are those which an editor should consider adding to the headline. In the example in Figure~\ref{fig:screenshot3}, the top three recommended keywords are already present in the headline; the two remaining recommendations, ``Irish Republic'' and ``GPO'', are both sensible suggestions for the article. 

In addition to the keyword recommendations, the system scores each headline for its ``shareability'' on two social media platforms: Twitter and Facebook; if the shareability score on either platform exceeds a threshold value, then an alert is displayed to the user. In the example in Figure~\ref{fig:screenshot3}, the article has exceeded the Facebook threshold but not the Twitter threshold, so only one of the two alerts is displayed. The newspaper's editor in charge of social media can use this information when deciding which stories should be posted and promoted on social media sites. The threshold is set to a relatively conservative value, so that most articles will not produce alerts, and only the most promising headlines will come to the editor's attention.

Ultimately, it is up to the editor to decide what action, if any, to take based on the information presented by the software. The editor has the leeway to add keywords in the headline in creative ways that fit the style of the story and the news organization, and she can also flexibly deal with any errors that may be produced by the keyword recommendation system, rather than blindly following its advice.


\section{Implementation}
\label{sec:Implementation}

The system consists of three major components: a user-interface front-end, a text analysis back-end, and a web server that mediates communication between the two. The user interface is implemented with HTML and Javascript and accessed via a web browser; its behavior is described and illustrated in the previous section. The web server is implemented in Python (based on the {\em Flask} framework), which allows easy integration with the text analytic back-end, which is also mainly implemented in Python. We use the {\em sklearn} module for regression and Stanford's {\em CoreNLP} Java suite for NLP \cite{Manning:2014}. The entire system is deployed on a web server and accessed by the client's web browser.


The back-end consists of two components---keyword analysis and shareability analysis---which operate independently of one another and are discussed in detail below.  

\subsection{Keyword Analysis}
\label{sec:Keyword}

The role of keyword analysis is to identify terms in the article body that are good candidates for inclusion in the headline. We believe that headlines containing informative and popular keywords can be both more appealing to readers and more prominent in users' search results and on news aggregator websites.

Processing of an input article begins with tokenization and named-entity recognition using {\em CoreNLP}, which identifies all entities (e.g. people, organizations, locations) in the article. Next, any known keywords appearing in the text are identified, by using a database of 90k keywords and their frequencies from Irish news articles in recent years, which we populated with data provided to us by two other Irish news-related projects \cite{Shi:2014,Bordea:2013}. This process results in a list of entities, which may be unique to the given article, and a list of keywords, which are known to have been encountered in previous news articles. These keywords and named entities are linked using a simple, rule-based approach that resolves pairs like ``Enda Kenny'' and ``(Mr.) Kenny'', yielding a single list of resolved keywords, along with a list of all positions in the text where each keyword appears.


Our keyword ranking system aims to capture the intuition that salient keywords should ideally be both {\em locally} prominent (i.e. appearing frequently in the given news article) and {\em globally} popular (i.e. appearing frequently in articles other then the current one). Thus, we calculate the weight $w$ of each keyword $k$ in the document $d$ as the weighted sum of its local weight $w_{local}$ and its global weight $w_{global}$:

\vspace{-1em} 

$$w(k,d) = \lambda\,w_{local}(k,d) + (1-\lambda)\,w_{global}(k)$$

The local weight is calculated as the normalized within-document frequency of the keyword, so that the most frequent keyword in the document gets a $w_{local}$ of 1. The global weight is calculated in a similar way, using the across-document frequencies from the keyword database and applying a nonlinear (log) transformation to compensate for the highly skewed distribution of these frequencies (note that it is possible for a keyword to have a zero global weight if it does not appear in our database; this is common for named entities in the article which have not been mentioned in the news before). The relative contributions of the local and global weights are balanced with the $\lambda$ parameter.

This formula was chosen as the simplest method (a linear combination) of combining the two factors. It is similar to a {\em tf-idf} score in that it combines both term frequency and document frequency, but it is critically different in that it rewards, rather than penalizes, terms that occur in many documents. This is a good thing because we believe that terms which may be very common (e.g. the names of well-known politicians or celebrities) can be good headline terms, and also because our method of selecting terms (via a closed set of keywords and automatic named entity detection) generally avoids selecting words which may be high-frequency but low-quality (like stopwords).

We manually set the value of $\lambda$ to achieve rankings which we subjectively deemed to be suitable.\footnote{We found that a value of 0.6 (i.e. slightly favoring local frequency over global frequency) worked well for our data, but this value changed depending on which keyword list we used. Ultimately, we combined both keyword lists, which introduced a large number of noisy terms. To suppress these noisy terms, we added an additional term to boost the score of keywords which were identified as named entities in the article (up to 0.2 of the overall weight).} This manual parameter setting allowed us to deploy our system quickly with acceptable performance, but a better option would be to learn these parameters automatically. To do so would require a dataset containing news articles, their headlines, and either some measure of the quality of the headline or an assurance that the headlines in the data set are ``good'', in order to 
guarantee that the parameters are set based on ``good'' headline examples. This type of data was not available to us when the system was under development.

This method ultimately assigns a weight to each keyword between 0 and 1.0, which determines its ranking in the analysis output (Figure~\ref{fig:screenshot3}). In the user interface, the weight is displayed in a table alongside the keyword's ``frequency'' and ``SEO Score'', which we consider to be more user-friendly than
 $w_{local}$ and $w_{global}$ themselves (the frequency is exactly the number of times the term appears in the article, and the SEO score is just $w_{global}$ scaled to the familiar scale of 0 to 100).

\subsection{Shareability Analysis}

%
%
%
%
%
%
%
%

The role of shareability analysis is to identify headlines that are likely to be shared on social media. With the rise of social media as dissemination channels for the news, headlines now need to be both informative and ``shareable''; that is, the headline somehow needs to attract people to post, share, and engage with the article on social media, in order to reach a large online audience.

According to the Reuters Institute Digital News Report~\cite{Reuters:2016}, Facebook and Twitter generate 54\% of the visits to online news sites, suggesting that direct visits to the home pages of news providers are being supplanted by social media mediated access. However, Facebook and Twitter are known to have quite different audiences and engage users in different ways ~\cite{Reuters:2015}. Users on Twitter generally actively search news and their consumption varies across news categories~\cite{Orellana-Rodriguez:2016:SNJ:2908131.2908154}, whereas on Facebook, news tends to be just encountered by sharing amongst friends. Therefore, in our system, we model the two social networks separately.

Using the Twitter streaming API we collected over 700k tweets and retweets posted by each one of 200 media outlets and journalist accounts for two time periods in 2013 and 2014, for a duration of 71 and 50 days, respectively. From the collected tweets we extracted all the URLs and used the Facebook and Twitter APIs to collect the number of times each URL was shared on Facebook or posted on Twitter. Because these posts were made by journalists, the links in the tweets are mainly to news articles, from which we extracted headlines. This step yielded a data set of 55k headlines with corresponding counts of social shares for each one.



We used a regression analysis to estimate the relationship between features of the headlines and the target variable of number of shares. Each headline in our collection is represented as a vector consisting of eight features covering three main aspects of the headline's content: the sentiment polarity (as computed by the {\em TextBlob} Python package), the presence of named entities, and the length in words. The complete list of features is presented in Table~\ref{tab:features}.

We used Regularized Linear Regression (RLR), Random Forest (RF) and Gradient Boosting Trees (GBT) as our methods for regression and used the metric Mean Squared Error (MSE) to assess their performance. We split our headlines set into 44k (80\%) for training and the remaining 11k (20\%) for testing. We train two different regression models, one for Facebook and one for Twitter. RF and GBT performed better than the RLR models. Between RF and GBT models, GBT performed slightly better than RF, although no significant difference was observed. On the basis of these results we use GBT as our method for regression. GBT have shown to outperform other models in classification and regression tasks and have been used successfully for audience engagement prediction~\cite{Diaz-Aviles:2014:PUE:2668067.2668072}. We observe that the models for Twitter and Facebook behave differently: comparing the values of the MSE for both models, predictions for shareable headlines on Facebook present an MSE of 41.8, while for Twitter the error is slightly smaller, 37.6.

Once the GBT models are trained, we store them and incorporate them into the system's pipeline. Every inputted headline receives two shareability scores, one for each social media site; however, in order to avoid triggering too many notifications to the journalist or news editor, the system only shows a result if the score is equal or larger than a manually-defined threshold
of 3.7 and 1.7 for Facebook and Twitter, respectively, which correspond to the median number of shares (on each platform) received by the headlines in our collection.

\begin{table}
	\centering
	\begin{tabular}{ll}
		\toprule
		\textbf{Feature} & \textbf{Description} \\ 
        \midrule
		neutral & \# of neutral sentiment words \\
		positive &\# of positive sentiment words \\
		negative &\# of negative sentiment words \\
		organizations & \# of {\sc Organization} entities \\ 
		persons & \# of {\sc Person} entities \\ 
		places & \# of {\sc Location} entities\\ 
		day & (T/F) headline contains the name of a day\\ 
		length & total \# of words in the headline\\ 
        \bottomrule
	\end{tabular}
	\caption{Headline features used for regression. The first six features are normalized by the length of the headline.}
	\label{tab:features}
\end{table}


\section{Evaluation}

The tool was developed in collaboration with \textit{}{The Irish Times}, and  several professional editors have tested its usability. The feedback from these sessions has been positive and has informed several design features. In particular, the color-coding and font-size features of the interface have been noted for their usability. On the basis of this success, we are now looking at integration into editors' daily workflow, to allow more usability data to be gathered.

Current tests of the system have identified some potential areas for improvement. The keyword system commonly fails to recognize when pairs of equivalent but non-identical keywords have the same referent; for example {\em Taioseach} and {\em Enda Kenny}, or {\em GPO} and {\em General Post Office}. While editors easily recognize this duplication, this error affects frequency counts, which in turn affect keyword rankings. This type of co-reference resolution is an open question in NLP research, with typical solutions relying on a rule-based or gazette-based approach to fix commonly-occurring cases.

The system could also be improved by moving from a static keyword database to a dynamic, real-time database. We were fortunate to be able to bootstrap our system with the keyword sources discussed in section~\ref{sec:Keyword}; however neither of these sources were created with this specific use-case in mind, and the static nature of these lists means that the keyword database will become outdated over time. Updating the keyword frequency counts on a rolling basis is an easy first step; but a more sophisticated approach is probably required, where new entities are added to the database over time, and more recent articles are given a greater weight than older articles. Because our system already identifies named entities in news articles, these entities could be added as new keywords in our database as they are encountered.

We are also evaluating the impact of using the tool on SEO, based on determining whether it improves article rankings in news aggregators and search engines. While the lack of click-through from Google News has led some to question its effectiveness at driving traffic to news sites \cite{Wauters:2010}, for \emph{The Irish Times'} website, Google News is a major source of referrals. An analysis of 30k \emph{Irish Times} articles (from 1/10/15 to 31/3/2016) has shown that articles listed on the Google News (Irish edition) front pages received significantly ($p<0.01$) more page views than unlisted articles; with Google News listed articles receiving almost twice as many views ($n=11,125$, $\mu=1665.5$ views per article) as unlisted articles ($n=19,339$, $\mu=892.4$ views per article). Google News' ranking algorithm is not publicly known, so the exact factors leading to this correlation are opaque; however, for practical purposes, if our keyword recommender leads to greater visibility on Google News, then we know it should increase readership.

Finally, the quality of our keyword recommendations can, in part, be assessed by noting whether the system's top-recommended keywords are already present in the original headline written for the article, as it shows that the system corresponds to human judgments (n.b., the keyword analysis only uses the article body, not the headline, as input). We processed a sample of roughly 3,000 Irish Times headlines with our system, and found that a majority of these (64\%) contained two or more of the top five keywords recommended by our system (in either the headline or the sub-headline), and a large majority (88\%) contained at least one of the recommended keywords. We take this as evidence that the keywords recommended by our system generally correspond with the types of keywords that a human editor would normally include in the headline. 

\section{Conclusion}

In this paper, we have presented a system for recommending keywords for inclusion in newspaper headlines and for identifying headlines with high potential shareability on social media. The system identifies plausible keywords that are both relevant to the given news article and popular overall in past news articles, in an effort to maximize both the reader interest and the SEO aspect of the headline. In addition, the system identifies headlines that are likely to receive above-average engagement on social media, allowing editors to effectively target their social media strategy. We believe that this tool can be a helpful component in modern, online-oriented newsrooms.


\section{Acknowledgments}

The authors would like to thank \emph{The Irish Times} for their funding and help on this project. This work is supported by Science Foundation Ireland through the Insight Centre for Data Analytics under grant number SFI/12/RC/2289.

\begingroup
	\bibliographystyle{named}
	\bibliography{ijcai16}
\endgroup

\end{document}